\RequirePackage{fix-cm}
\documentclass[runningheads]{llncs}
\usepackage{graphicx}
\usepackage{cite}
\begin{document}

\title{DocReader: Bounding-Box Free Training of a Document Information Extraction Model}

\titlerunning{DocReader}        

\author{Shachar Klaiman \inst{1} \and Marius Lehne \inst{2}}


\institute{SAP AI \\
            Dietmar-Hopp-Allee 16, 69190 Walldorf, Germany \\
            \email{shachar.klaiman@sap.com}          
\and        SAP AI \\
            Münzstraße 15, 10178 Berlin, Germany\\
            \email{marius.lehne@sap.com}
}

\date{Received: \today / Accepted: \today}

\maketitle

\begin{abstract}
Information extraction from documents is a ubiquitous first step in many business applications. During this step, the entries of various fields must first be read from the images of scanned documents before being further processed and inserted into the corresponding databases. While many different methods have been developed over the past years in order to automate the above extraction step, they all share the requirement of bounding-box or text segment annotations of their training documents. In this work we present DocReader, an end-to-end neural-network-based information extraction solution which can be trained using \textit{solely} the images and the target values that need to be read. The DocReader can thus leverage existing historical extraction data, completely eliminating the need for any additional annotations beyond what is naturally available in existing human-operated service centres. We demonstrate that the DocReader can reach and surpass other methods which require bounding-boxes for training, as well as provide a clear path for continual learning during its deployment in production. 
\keywords{Document Information Extraction \and Deep Learning \and OCR \and Attention \and RNN}
\end{abstract}

\section{Introduction}
\label{sec:intro}

Information extraction from documents is an indispensable task in many scenarios. The information extracted can vary depending on the down stream task, but normally includes \textit{global} document information which is expected to appear once on the document, e.g., document date, recipient name, etc.,  and \textit{tabular} information which can be in the form of an actual table or an itemized list in the document. In the context of business documents, where we will focus our discussion on, global information is also referred to as header fields whereas tabular information is often referred to as line-items. There is a myriad of different document templates which normally prevents one, in all but very narrow applications, from easily developing a rule-based extraction solution. In some large-scale applications, centralized spend management for example, one could even reach the extreme but realistic situation where most templates are seen only once. 

The majority of current state-of-the-art automated information extraction solutions are based on an initial optical character recognition (OCR) step. As such they follow the generic two-step pipeline: first extract all the text from the document and only then localize and retrieve the requested information from the previously extracted text \cite{zhang2020trie, xu2020layoutlm}. There have been multiple approaches in the literature for solving the localization step. While some initial approaches were ``pure'' natural language processing (NLP) approaches, e.g., \cite{lample2016neural}, studies demonstrated the importance of including also the positional information of the text on the image. This led to a different set of models incorporating methods from both computer vision (CV) and NLP, see for example \cite{katti2018chargrid}. The described two-step setup, inevitably requires position-labels, i.e., bounding-box annotations, for training the models. 

Training the localization models requires annotating tens-of-thousands of records, leading to substantial costs as well as a significant delay in training before the annotation work is over. The training dataset contains a large set of pairs of documents and labels. The labels which are constructed using the annotation results are a collection of the fields which we want to extract from the document, i.e., the various textual elements we wish to extract from the document, as well as bounding boxes for each of these fields, i.e., the coordinates of a rectangle surrounding each of the text snippets we would like to extract from the document. Yet another unavoidable complexity with the annotation process, lies in the need to select samples from the historical database to be annotated. Bad sampling leads to multiple annotation rounds which can further delay achieving the needed model performance. The sampling is usually needed due to the shear size of the historical database in many cases.

In this work we present a different approach to information extraction models. Our approach is motivated through the vast amounts of historical manually extracted data already available wherever a human-based extraction task is being performed. In contrast to the training data used for the localization models discussed above, this historical data does not contain any bounding box information. Thus, we only have access to the document images and the extracted key-value pairs. We therefore need to design an information extraction model which can be trained without any bounding-box annotations. This led us to the design of the DocReader model which is trained using only the \textit{weakly} annotated data, i.e., training data consisting of only images and key-value string pairs, but can reach and surpass the performance of models using \textit{strongly} annotated data. The DocReader model demonstrates how a data-driven approach, where the model is designed based on the available \textit{very large} training data rather than preparing the training data to fit the model, can alleviate many of the challenges in acquiring training data while still providing state-of-the-art performance on the given task. 

\begin{figure}[htp]
    \centering
    \includegraphics[width=0.75\textwidth]{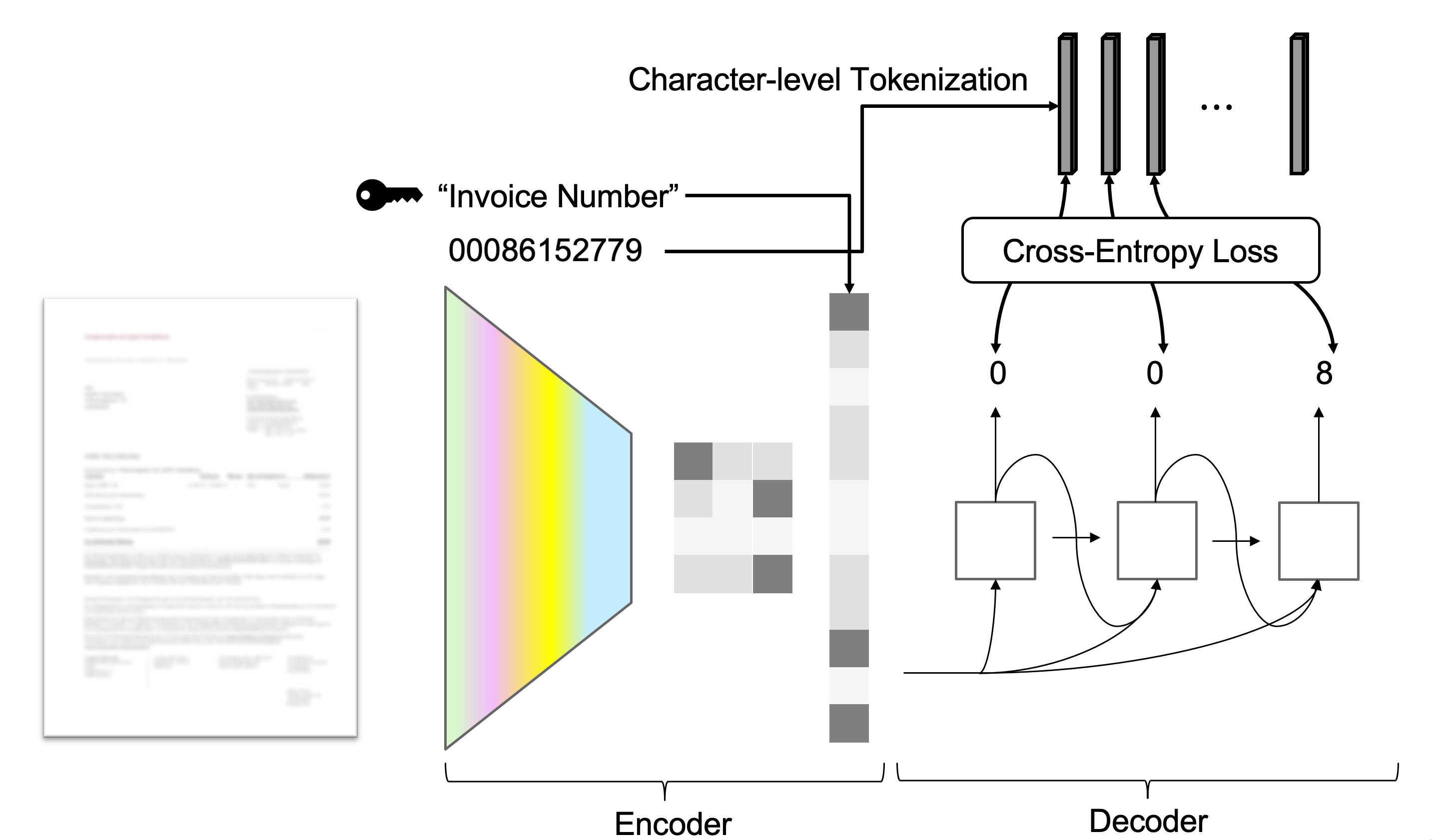}
    \caption{Overview of the complete DocumentReader model. A document and an extraction key are provided as an input. The key is used to condition the attention layer. The image is first processed by an encoder. The resulting feature map is passed together with the key into an RNN-based decoder. The cross-entropy loss is computed from character based tokenization of the ground truth and the projected output of the RNN.}
    \label{fig:model}
\end{figure}

\section{Related Work}

The standard approach to information extraction is a two stage process, that requires an initial OCR step followed by a second information localization step. Localization can be performed on the extracted text sequence, by training a NER model \cite{gralinski2020kleister}. With such an approach however spatial information is lost. In recent publications different methodologies are proposed to incorporate this spatial or relational information to improve extraction performance. In NER a simple way is to add coordinates to the word tokens \cite{sage2020end} or augment the word tokens with positional embeddings \cite{xu2020layoutlm}. In \cite{liu2019graph} the authors augment each text token with graph embeddings that represents the context of that token.  In \cite{majumder2020representation} the authors propose a system that identifies a set of candidates for a query based on a pretrained generic NER model. Scoring of the candidates is performed by an embedding of its neighborhood. The flexibility of this approach is limited by the availability of an external NER model that has a schema that generalizes the schema of the queries. 

CharGrid \cite{katti2018chargrid} combines extracted text with methodologies from computer vision. The extraction task is formulated here as an instance segmentation task with a two-dimensional grid of encoded characters as input. The predicted segmentation masks and bounding boxes are overlaid with the word-boxes of the OCR in order to retrieve the relevant text from the invoice. BertGrid \cite{denk2019bertgrid} extended the CharGrid input concept by using a grid of contextualized embeddings as an input instead of encoding single character tokens. 

All of the above systems rely on an external OCR solution. This means that the error of the extractions is bounded by the OCR error, meaning that even a perfect extraction model would still be limited by the OCR accuracy. Additionally, strong annotations such as token label or bounding boxes are required.

Another relevant line of research deals with the OCR in the wild task. Here one wishes to extract the text from an image, but the text is normally very short and embedded in some other part of the image. Since the task requires the extraction of all the text on the image, models developed to solve this task can be trained with only text labels and do not require location information through bounding boxes. The authors in \cite{wojna2017attention} introduce an end-to-end OCR system based on an attention mechanisms to extract relevant text from an image. Similarly, in \cite{bartz2018see}, the authors use spatial transformers to find and extract from text from relevant regions. While these models can be trained without bounding-box information, they have been so far limited to extracted all the text from the image and as such could not be directly ported to the document information extraction task. 

\begin{figure}[htp]
    \centering
    \includegraphics[width=\textwidth]{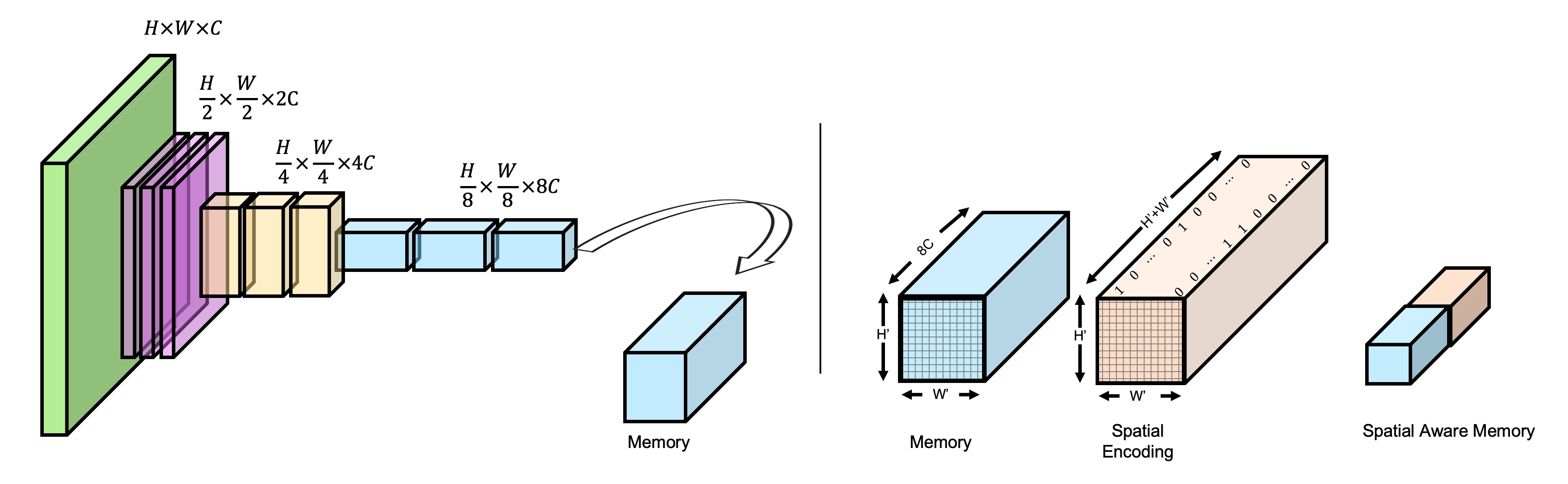}
    \caption{Left: Structure of the encoder model. An input image is passed through several convolution blocks. Right: Spatial augmentation of the memory. Coordinate positions are separately encoded as one-hot vectors and concatenated to the memory on the channel axis.}
    \label{fig:encoder}
\end{figure}

\section{Method}
\label{sec:method}

\subsection{Model Architecture}
\label{sec:model}

We propose an end-to-end neural network architecture that generates a specific text from an input image specified by an extraction key, e.g. invoice number on an invoice. This architecture, as sketched in Figure \ref{fig:model}, follows an encoder-decoder pattern, where the feature map coming out of the encoder is conditioned on the provided extraction-key. 

The input to the model is an image of height $H$ and width $W$. It is passed into an encoder with a VGG-inspired \cite{simonyan2014very} structure identical to \cite{reisswig2019chargrid}. A sketch of the encoder is shown in Figure \ref{fig:encoder} on the left. It consists out of an initial convolution layer with a base channel size of $C$ followed by three convolution blocks. Each block is comprised of a sequence of three convolution layer. The first layer in each of the first three blocks are convolutions with a stride of $2$. Hence, the resolution after the first three blocks is reduced by a factor of $8$. The number of channels doubles with each block. The convolutions in block 3 are dilated with a factor of 2. Every convolution layer is followed by a dropout layer, a batch-normalization layer, and a ReLu activation function. The output of the encoder is a dense intermediate representation of the input image with the dimensions $(H^\prime, W^\prime, 8C)$, which is also often referred to as the \textit{memory}. This memory is then used by the decoder to generate the output sequence.

Before passing the memory into the decoder we fuse the output of a spatial-encoder along the channel axis of the memory, thereby adding positional information to each of the features in the memory. The spatial encoder could be a straight-forward one-hot encoding of the height and width of the memory, as depicted in Figure \ref{fig:encoder} on the right, and first presented in \cite{wojna2017attention}, a fixed one-dimensional encoding \cite{vaswani2017attention} or a trainable positional embedding. In our model we have experimented with all of the above options and concluded that the one-hot positional encoding provides the best compromise between accuracy and model size. As a result we obtain a spatially augmented memory.

\begin{figure}[htp]
    \centering
    \includegraphics[width=\textwidth]{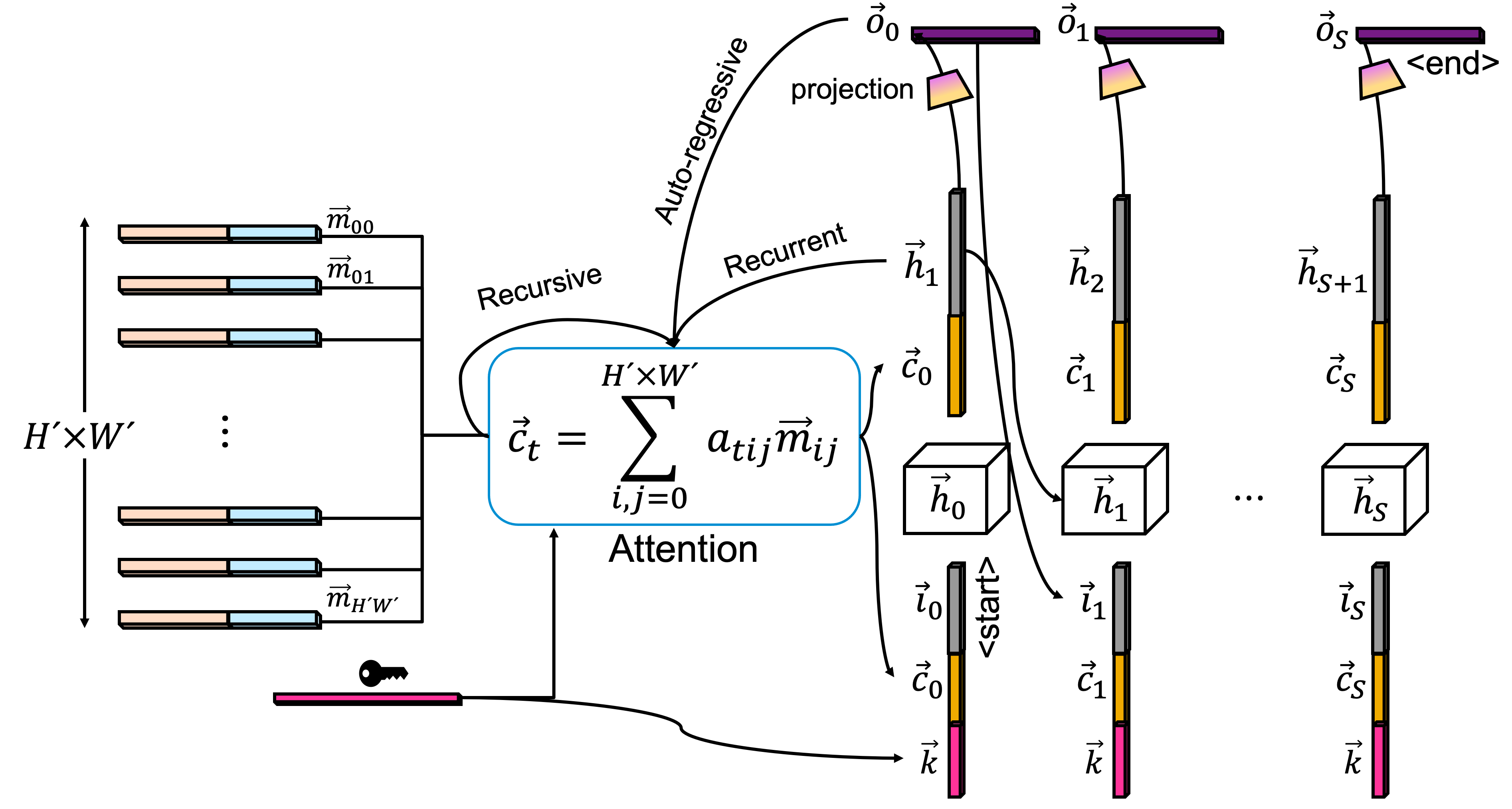}
    \caption{Network structure of the decoder. An attention layer computes of the memory and the key a context vector. In each RNN step the cell receives the key, context vector and the previous character. The state of the RNN is concatenated with the context vector and passed into a projection layer that outputs a character.}
    \label{fig:decoder}
\end{figure}

The decoder, see Figure \ref{fig:decoder}, is based on a recurrent neural network coupled with an attention layer. It receives two inputs. The first input is the spatially augmented memory. The second input is the key that determines the field of interest to be extract from the document, e.g., invoice number. This key is encoded using either a one-hot or a trainable embedding. For the recurrent layer in the DocReader we chose a LSTM \cite{hochreiter1997long} and for the attention layer we use an \textit{augmented} and \textit{conditioned} version of the sum-attention layer \cite{bahdanau2014neural}. 

Documents normally contain many potential relevant text sections in various degrees of readability. Localizing and recognizing the correct information on the document is thus a key challenge in the information extraction task. On a coarser level the correct location of the information of interest needs to be identified. On a more detailed level the correct sequence characters need to be pinpointed in each step. To tackle these challenges, we augment the base sum-attention layer with additional inputs as depicted in Figure \ref{fig:decoder} as explained in detail below. 

The attention score is computed at each decoding step for each spatial element in the memory, represented by the vector $\vec{m}_{ij}$. We additionally condition the attention layer on the input key by adding the vector representation of the key $\vec{k}$ as an additional attribute to the scoring function. Making the model aware of the key allows us to train a model on multiple keys simultaneously. We further introduce two additional modifications to the attention layer for supporting the character by character decoding by providing additional information about the previous decoding step. We include the previously predicted character $\vec{o}_{t-1}$ into the scoring function. We refer to this attribute of the attention as auto-regressive. We also include the previous step's "flattened" attention weights, $\vec{a}_{t-1}$, as an additional parameter to the scoring function. With this additional information, the attention layer can use the attended location of previous decoding step to assist it in deciding where to focus next. We call this mechanism recursive attention. This idea is similar to a coverage vector that has been used in neural machine translation and was introduced in \cite{tu2016modeling}. 

In summary, we provide the scoring function with the memory $\vec{m}_{ij}$ and the current LSTM cell state $\vec{h}$, the embedded key $\vec{k}$, the previous attention weights $\vec{a}_{t-1}$ and the previous predicted character $\vec{o}_{t-1}$. The scoring function then reads:
\begin{equation}
\label{eq::att1}
f\left(\vec{m}, \vec{k}, \vec{a},\vec{b},\vec{c} \right) = \tanh\left(\vec{W}_m\vec{m} + \vec{W}_k\vec{k}  + \vec{W}_a\vec{a}  + \vec{W}_b\vec{b}  + \vec{W}_c\vec{c}\right)
\end{equation}
\begin{equation}
\label{eq::att2}
    a_{tij} = softmax\left(\vec{v}^t f\left(\vec{m}_{ij}, \vec{k}, \vec{h}_{t-1},\vec{o}_{t-1},\vec{a}_{t-1} \right)\right)
\end{equation}
The context vector is then calculated from the memory in the following way: 
\begin{equation}
\label{eq::att3}
    \vec{c}_t =\sum_{i,j}^{H^\prime \times W^\prime} a_{tij}\vec{m}_{ij}
\end{equation}

The attention layer outputs the context vector for the next step as well as the attention weights themselves. The resulting context vector is then concatenated to the character embedding $\vec{l}_{t-1}$ of the input at time $t-1$ and the key embedding to form the LSTM cell input. 

\begin{equation}
    \vec{h}_t = \textrm{RNNStep} \left( \vec{h}_{t-1}, \vec{l}_{t-1}, \vec{c}_{t}, \vec{k} \right)
\end{equation}

The LSTM cell output at each step is concatenated with the step’s context-vector and fed into a softmax-layer which projects the input of the layer into the prediction probabilities over the different characters in the vocabulary. 


\subsection{Training}
\label{sec:training}

As an end-to-end solution, one would expect to train the DocReader model from scratch. Indeed, for other models with similar architectures, e.g., attentionOCR \cite{wojna2017attention}, the authors have shown that given sufficient training data, one can train the entire network from scratch. In contrast to the "OCR in the wild" task studied in \cite{wojna2017attention}, we focus here on document information extraction. The main difference is that we are not interested in all the text in the image but rather only a very specific segment of text from the image. This makes learning from scratch substantially more complicated. Characters from the expected string could appear in multiple places on the document and the task of initially localizing the information on the image is much harder. Furthermore, given a randomly initialized encoder, the feature map going into the decoder has yet to learn the concept of characters which means we are trying to simultaneously learn to "see", "find",  and "read" which is expectedly hard. We emphasize, that the above challenges in the document extraction task compared to the "OCR in the wild" task apply in general and are not limited to the difficulty of training the model from scratch. The model now extracts only a specific text element from the image and not the entire image text making the \textit{signal} the model needs to learn extremely sparse in comparison. 

The above challenges, led us to consider transfer-learning as a starting point for training the DocReader. Since the input to the DocReader is an image, it is natural to consider any state-of-the-art pretrained image network. This approach was used in multiple previous studies \cite{wojna2017attention, hebrewsigns}. Since the DocReader not only needs to distinguish textual from non textual areas in the image but rather also distinguish between different texts in the image, a generic image classifier would not be a good choice here. Therefore, we chose here, to use the encoder of a pretrained OCR model instead. Specifically, we use the encoder from the recently published OCR model: CharGrid-OCR \cite{reisswig2019chargrid}. The usage of a pretrained OCR model, provides an excellent starting point for the DocReader, since the feature map obtained from the decoder already allows the decoder to distinguish between different characters in the image. We note in passing, that our choice to facilitate the training using a pretrained OCR model does not in principle exclude the possibility of training the DocReader from scratch given enough training data and computational resources. 

In practice, the DocReader is trained using a two phase procedure. During the first phase the weights of the encoder, which are taken from the pretrained OCR model, are frozen. The second phase commences after convergence is reached in the first training phase. In the second phase all weights in the model are trainable.  As explained above, the decoder must "see", "find", and "read" from the encoder output. By first freezing the encoder, we allow the decoder to concentrate on the latter two actions instead of trying to optimize all three at once. Our experiments show that without this initial training phase, the decoder tends to quickly overfit the training data, completely bypassing the attention module. In other words, the decoder "memorizes" rather than "reads". As we shall explicitly demonstrate in a later section, the results of the DocReader after only the first training phase are still very poor. Thus, we cannot leave the encoder fixed throughout and competitive results can only be achieved by fine-tuning the entire network. 

Normally, one wishes to extract multiple, possibly very different, fields from the input document. With respect to the DocReader model's training, this raises the question whether one should train multiple single-key models or rather a single multi-key model. The DocReader's built-in key-conditioning allows one to do both. In practice we observe that multi-key models tend to perform better than many single-key models. The improvement likely stems from better training of the encoder as the multiple-keys, which comprise of different characters and appear on different locations in the image, allow for a better coverage of the feature map being trained. Nevertheless, one must also consider the issue of data sampling and training time when choosing how many keys to simultaneously train on. Since the DocReader extracts in every step the value for a single-key, the number of records in the training set scales linearly with the number of keys being trained on, because we add the same invoice with a different label to the training set. Furthermore, some of the keys do not appear on all documents which means the model needs to learn to predict a missing value for certain document and key pairs. We eventually divide the keys into groups having similar missing-value support in the training set. This allowed us to avoid predicting to many false-negatives on keys which are almost always present but avoid the need to exclude more rare fields.

\section{Information Extraction From Invoices}
\label{sec:Results}

\subsection{Data}
\label{sec:data}

Our training dataset consists of $~1.5M$ scanned single-page invoices. The invoices are in different languages and the majority of invoice templates appear only a single time in the dataset. Each invoice image is accompanied with the human extracted values (strings) for a set of fields, keys, for example, invoice number, invoice date, etc. Our test set which comprises of 1172 documents from 12 countries, contains also multi-page documents. By excluding overlapping vendors between the training and test sets, we make sure that the test set does not contain any of the invoice templates from the training set. We note that the above strict separation of templates between training and test set produces a very stringent performance measure, but provides a very realistic measure of the generalization capabilities of the trained model. 

In order to compare with other state-of-the-art approaches which require bounding-box annotations for training, we also fully annotated a training set of $~40k$ scanned invoices, i.e., key-value strings and bounding-box annotations. The invoices in this set are in multiple languages with the very large variety in templates. We shall refer to this training set as the strongly-annotated training set in the following. Also the vendors of the strongly-annotated set are excluded from the above test set.

Both of the above datasets are proprietary in-house datasets. Therefore, in order to visualize the DocReader's extractions we use documents from the invoice category of the RVL-CDIP test set \cite{harley2015icdar}. 

\subsection{Implementation Details}
\label{sec:implementation_details}
As described above, we reuse a part of the encoder of the pretrained CharGrid-OCR model \cite{reisswig2019chargrid} as the encoder of the DocReader. Here, however, lies one of the big challenges in training the DocReader without any positional information, i.e., bounding boxes. The OCR model is usually trained on crops from the full, 300 dpi, document. Since the OCR training data contains positional information, cropping is not an issue. When we now want to use the same encoder for the DocReader, however, we no longer have any positional information and therefore we cannot crop the training images. In other words, we must feed the DocReader the entire image of the document. As a result of this, we must reduce the resolution of the scanned image, and thus pretrain the OCR model on the reduced resolution as well. In practice we found that an input resolution of $(832, 640)$ works very well. 

We choose the base channel number $C$ to be $32$. For the convolutional layer in the encoder we use $3 \times 3$ kernel. The dropout layer in the encoder have a dropout probability of $0.1$ during training. The LSTM cell has a size of $256$ and the attention layer has a dimensionality of $256$. The token vocabulary encompasses ASCII characters including punctuation characters\footnote{This is equivalent with the set of characters returned by string.printable in Python}. We also add special tokens for start of sequence, padding, end of sequence and out of vocabulary. For training we use a batch size of $16$.

\begin{figure}[ht]
    \centering
    \includegraphics[width=\textwidth]{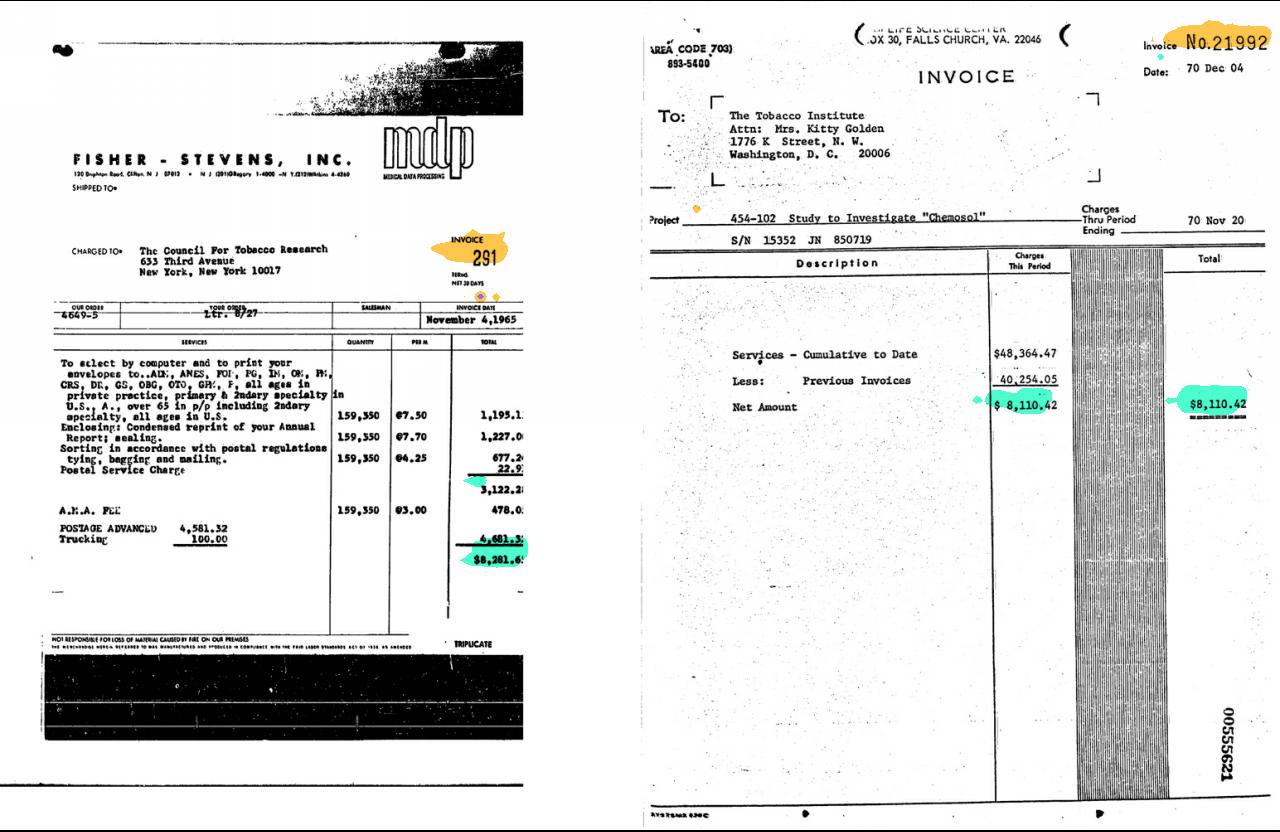}
    \caption{DocReader attention maps for the predictions of invoice number (orange) and invoice amount (turquoise) on the two invoice scans taken from the invoice category of the RVL-CDIP test set \cite{harley2015icdar}. When multiple instance of a field appear on the invoice, the DocReader attends to all instance simultaneously.}
    \label{fig:number_amount}
\end{figure}

The ground truth is provided as a sequence of characters. We add a start token at the beginning of the training and an end token at the end of the sequence. We observe that adding several warm-up tokens at the beginning of the sequence can facilitate the decoding process. This provides the attention layer some extra steps to determine the correct location of the text to be extracted. During training we make use of teacher forcing. For inference we feed the output of the previous step into the decoder. We stop decoding if either a stop token was encountered or the maximum sequence length has been reached. We compute a cross-entropy loss for each character individually and sum the loss over the sequence. 

Special care must be taken when choosing the learning rate for training the DocReader. As explained in Section \ref{sec:model}, the DocReader model is constructed from 3 modules: the encoder, the attention, and the decoder. If the learning rate is too high, the attention module will simply be ignored and the decoder will memorize the input samples and will not generalize. One can see this very clearly from the attention maps where the attention in such a case is spread over uniformly over the entire image instead of targeting the desired key. For our training we use the Adam optimizer and a learning rate of $2\cdot10^{-4}$. We note in passing that automated \textit{optimal} learning rate algorithms such as that detailed in \cite{smith2017cyclical} fail to resolve this type of dependency since the training loss will continue to go down even if the attention module is \textit{bypassed}.

\begin{figure}[h]
    \centering
    \includegraphics[width=\textwidth]{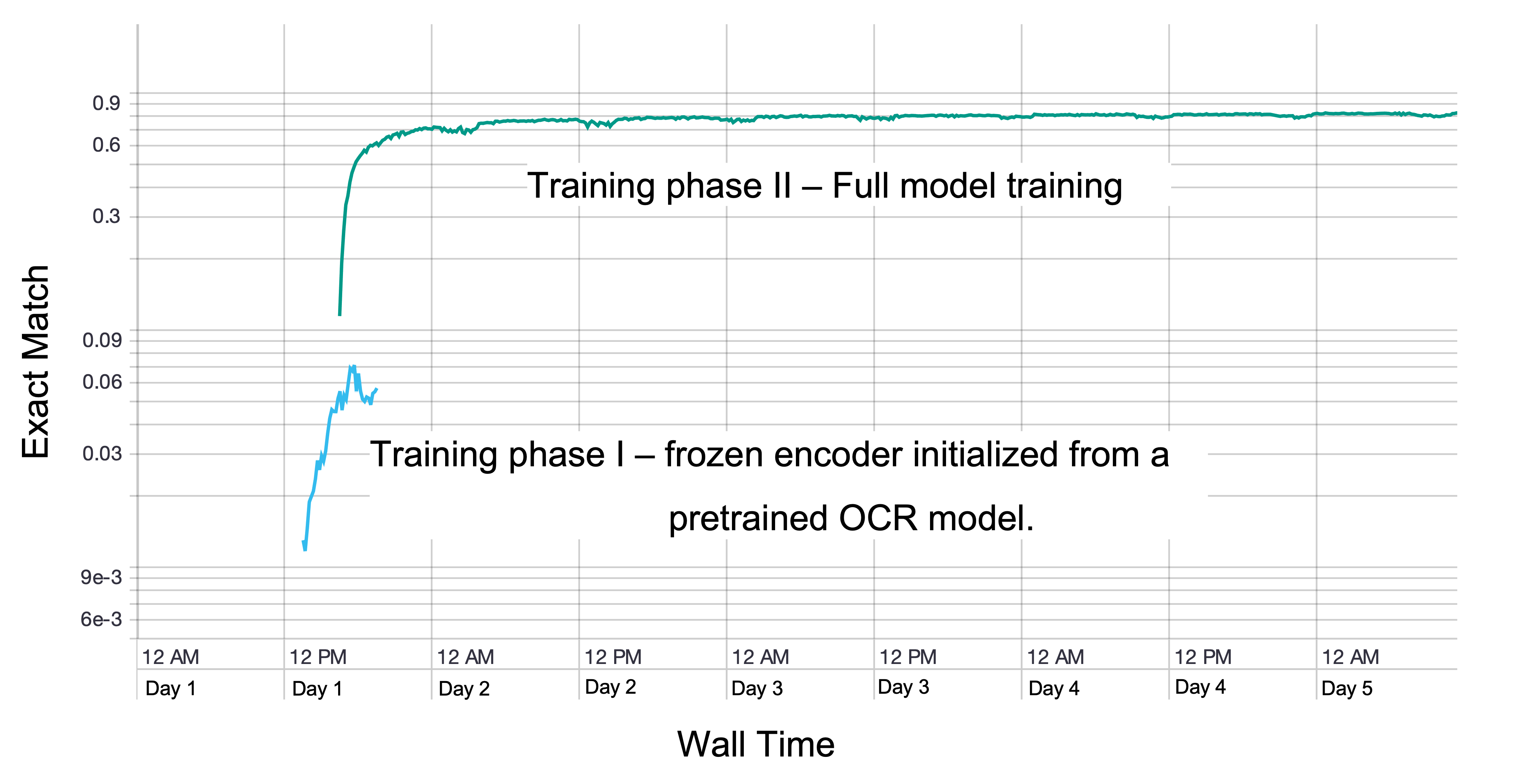}
    \caption{Exact string match metric on the evaluation set during the two phase training regimen as described in Section \ref{sec:training}. During the first phase here shown in blue the encoder weights are initialized from a pre-trained OCR model \cite{reisswig2019chargrid}, but remain frozen. In the second phase the training is continued on all weights of the model.}
    \label{fig:training_phases}
\end{figure}

\subsection{Experiments and Results}
\label{sec:experiments_and_results}

We trained and evaluated our model on the extraction task of 6 different fields from invoice scans. Motivated by the ratio of documents where the different fields occurs we trained 3 different models. The first was trained to extract the fields: invoice number, invoice date, total amount, and currency. The second model was trained on extracting the purchase order number (PO) from invoices. The third model was trained to extract the tax amount from the invoice. 

Training is performed in the two phase training procedure as described in Section \ref{sec:training}. The model performance during training time on the evaluation set is shown in Figure \ref{fig:training_phases}. We observe an increase of model performance in the first phase that levels off at at quite low value. Hence, using only the features from the frozen pre-trained encoder is not enough to perform well on this task. Fine-tuning the whole model enables us then to achieve significantly better results.

The model's performance was evaluated using an exact string match for the predicted values against the ground truth values. Fields that are not present on a document are represented by empty strings. The metric therefore also covers cases where the model falsely extracts a value for a non-existing fields or where the model falsely does not return a value for an existing field, i.e., false negatives and false positives. We chose this stricter metric since it guarantees automation of the extraction process. It is worth keeping in mind that, in contrast to the DocReader, models which copy the predicted value from an OCR output depend on the correctness of the OCR which cannot be controlled by the extraction model. 

\begin{table}
\caption{Exact string match of the predicted and ground-truth strings on the test set defined in Sec. \ref{sec:data}. We compare the DocReader trained on 1.5M invoices to the state-of-the-art CharGrid trained on 40k invoices \cite{katti2018chargrid} on the extraction task of 6 different fields from invoice scans. For predicting the invoice currency, the displayed CharGrid accuracy is the combined accuracy of CharGrid and a separately trained OCR-text currency Classifier. See the text for more details. For four of the six fields, we also show the performance of the DocReader model when trained using only the training set used to train the CharGrid model (DocReader(40k)). }
\label{tab:1}       
\begin{center}
\begin{tabular*}{\textwidth}{@{\extracolsep{\fill}} lrrrrrr }
\hline\noalign{\smallskip}
Model & Number & Date & Amount & PO & Tax & Currency\\
\noalign{\smallskip}\hline\noalign{\smallskip}
CharGrid & 0.72 & 0.82 & 0.87 & \textbf{0.82} & 0.47 & $0.82^{*}$\\
DocReader & \textbf{0.79} & \textbf{0.83} & \textbf{0.90} & 0.79 & \textbf{0.80} & \textbf{0.97}\\
DocReader(40k) & 0.58 & 0.78 & 0.84 & -- & -- & 0.89\\
\noalign{\smallskip}\hline
\end{tabular*}
\end{center}
\end{table}

In Table \ref{tab:1} we compare the DocReader's performance on all the extracted fields. We compare the performance to a CharGrid model \cite{katti2018chargrid} which was trained on the $40k$ fully annotated set, see Sec. \ref{sec:data}. The CharGrid model was previously shown to provide state-of-the-art results on invoice information extraction. The DocReader model show comparable results on the invoice date, invoice amount, and purchase order number fields, and substantial improvement on invoice number, tax amount, and currency. 

Figure \ref{fig:number_amount} depicts the attention maps for the invoice number and invoice amount predictions of the DocReader model. The attention maps are created by summing the attention maps of all single character prediction steps and then resizing the attention map to the original image size. The two invoices shown are taken from the invoice category of the RVL-CDIP test set \cite{harley2015icdar}. The invoices in this test set are of rather poor scanning quality and appear rather different than the invoices the model was trained on. Still, the DocReader succeeds in extracting the chosen fields accurately. It is interesting to note, that in cases of multiple instance of the field on the invoice, the DocReader simultaneously reads from all the instances.  

\begin{figure}[h!]
    \centering
    \includegraphics[width=\textwidth]{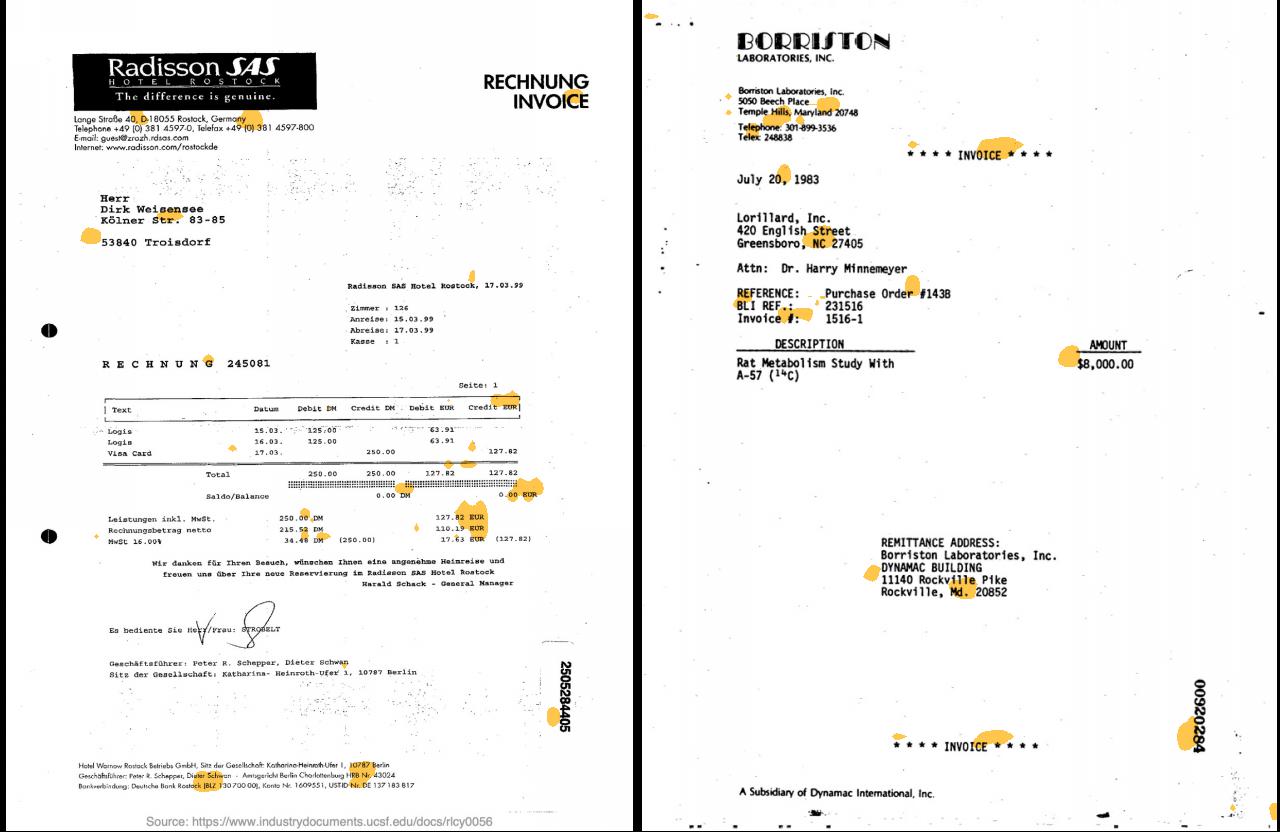}
    \caption{DocReader attention maps for the predictions of invoice currency. If the currency symbol/identifier is directly on the invoice, the DocReader attends to that identifier directly. In cases where the symbol is not on the invoice or where the symbol is not sufficient to make a currency prediction, e.g., the \textdollar{} sign is not sufficient to distinguish between USD and AUD, the DocReader infers the currency from the address section of the invoice as a human extractor would. The two invoices were taken from the invoice category of the RVL-CDIP test set \cite{harley2015icdar}.}
    \label{fig:currency}
\end{figure}

Compared to OCR-based extraction models, the DocReader can also predict \textit{global} fields which cannot always be found in the text on the invoice itself. One such field, for example, is the invoice currency. In many invoices there is no currency symbol at all. Even when a currency symbol is on the invoice, it is often not sufficient to make a concrete prediction. Consider a US invoice where the amount has the \textdollar{} symbol next to it. One cannot infer that the currency is USD from the currency symbol. Instead, one would have to also examine the invoice address fields to determine the actual invoice currency. Since the DocReader is not restricted to localize first and then copy the value from the OCR output, it can directly infer the currency from the invoice image. As depicted in Figure \ref{fig:currency}, in cases where the currency symbol is on the invoice, the DocReader reads the symbol directly whereas whenever the currency symbol is insufficient to make a currency prediction, the model attends the address field as well in order to determine the correct currency. Our previous state-of-the-art solution for currency employed a combination of the OCR-based CharGrid model with a fallback on a OCR-text-based currency classifier in the case the CharGrid could not retrieve any decisive currency prediction. As Table \ref{tab:1} clearly shows, the DocReader offers a very significant improvement on the performance on this \textit{global} field.

\section{Discussion}

The results of the experiments described in Section \ref{sec:Results}, demonstrate how in situations where we have a very large amount of \textit{weakly} annotated historical data, the DocReader model can yield state-of-the-art results while freeing us from the need to additionally annotate any data. As can be expected (see Table \ref{tab:1}), without access to a substantial amount of historical data, methods based on positional annotations, i.e., \textit{strong} annotations, have a clear advantage over the DocReader. 

One point to keep in mind when using the historical extraction data directly, is that the model is only able to learn the extraction pattern which already exists in the data. Thus, one cannot define a set of desired extraction rules which other approaches would give as their annotation guidelines. Instead, we can at most try and restrict certain rules by preprocessing the values but in principle the model learns to read the extraction pattern set by the humans performing these task before. This could also be perceived as an inherit advantage of the method since we do not need to understand the extraction logic before hand, the model will learn this logic by itself. It does though require that previous extractions have been made in a consistent manner.

We note that the DocReader model can be thought of as a very generic targeted OCR model. The flexible conditioning mechanism we incorporated directly into the attention module allows one to choose essentially any consistently appearing information on an image and read it out. The model also provides automatic post-processing so long as the post-processed values are consistent across the training set. Thus one can, for example, directly extract the date in ISO format even when the training set is composed from documents from many countries where the date format on the documents themselves can be very different.

Since we do use a trained OCR model to initialize the DocReader encoder, it is fair to ask whether we are indeed completely free of positional annotations as these are used in the training of the OCR model. As we mention in Section \ref{sec:training}, this initialization of the model might not be strictly necessary \cite{wojna2017attention}. Furthermore, the annotations for training the OCR model are generated and require no manual annotations as explained in \cite{reisswig2019chargrid}. This means that even with the DocReader initialization from the OCR model, no additional manual positional annotations are required at all.

\section{Conclusion}

The DocReader model, presented in this work shows how a data-driven model design can lead to substantially improved performance on the document information extraction task. By using only the \textit{available} historical extraction data, the DocReader mitigates the challenges and costs which come with specialized data annotations. Our end-to-end approach allows us to go beyond the standard extraction tasks of identifying the correct content and reading the corresponding value. The DocReader model can also directly solve image classification tasks, e.g., predicting invoice currency, making use of very diverse information available on the document image such as the document template and the interplay between different fields.


\bibliographystyle{spmpsci_unsort}      

\bibliography{bibliography.bib}   

\end{document}